\documentclass[10pt,twocolumn,letterpaper]{article}

\usepackage{cvpr}
\usepackage{times}
\usepackage{epsfig}
\usepackage{graphicx}
\usepackage{amsmath}
\usepackage{amssymb}
\usepackage{subcaption}

\usepackage{multirow}
\usepackage{booktabs}
\usepackage{caption}

\usepackage{tabularx}
\newcolumntype{L}[1]{>{\raggedright\arraybackslash\hspace{0pt}}p{#1}}
\newcolumntype{P}[1]{>{\centering\arraybackslash\hspace{0pt}}p{#1}}
\newcolumntype{C}[1]{>{\centering\arraybackslash\hspace{0pt}}m{#1}}
\newcolumntype{R}[1]{>{\raggedleft\arraybackslash\hspace{0pt}}p{#1}}

\def\g[#1]{\color[gray]{#1}}

\usepackage[pagebackref=true,breaklinks=true,letterpaper=true,colorlinks,bookmarks=false]{hyperref}

\cvprfinalcopy 

\def\cvprPaperID{110} 

\ifcvprfinal\fi
\begin{document}

\title{Dual Attention Networks for Multimodal Reasoning and Matching}

\author{Hyeonseob Nam\\
Search Solutions Inc.\\
{\tt\small hyeonseob.nam@navercorp.com}
\and
Jung-Woo Ha\\
NAVER Corp.\\
{\tt\small jungwoo.ha@navercorp.com}
\and
Jeonghee Kim\\
NAVER LABS Corp.\\
{\tt\small jeonghee.kim@naverlabs.com}
}

\maketitle

\begin{abstract}
We propose Dual Attention Networks (DANs) which jointly leverage visual and textual attention mechanisms to capture fine-grained interplay between vision and language.
DANs attend to specific regions in images and words in text through multiple steps and gather essential information from both modalities.
Based on this framework, we introduce two types of DANs for multimodal reasoning and matching, respectively.
The reasoning model allows visual and textual attentions to steer each other during collaborative inference, which is useful for tasks such as Visual Question Answering (VQA).
In addition, the matching model exploits the two attention mechanisms to estimate the similarity between images and sentences by focusing on their shared semantics.
Our extensive experiments validate the effectiveness of DANs in combining vision and language, achieving the state-of-the-art performance on public benchmarks for VQA and image-text matching.

\end{abstract}

\section{Introduction}
Vision and language are two central parts of human intelligence to understand the real world.
They are also fundamental components in achieving artificial intelligence, and a tremendous amount of research has been done for decades in each area.
Recently, dramatic advances in deep learning have broken the boundaries between vision and language, drawing growing interest in their intersection, such as visual question answering (VQA)~\cite{antol2015vqa,zhou2015simple,noh2016image,yang2016stacked}, image captioning~\cite{xu2015show,annehendricks2016deep}, image-text matching~\cite{hodosh2013framing,karpathy2014deep,mao2015deep,wang2016learning}, visual grounding~\cite{plummer2015flickr30k,hu2016natural}, etc.

One of the recent advances in neural networks is the attention mechanism~\cite{mnih2014recurrent,bahdanau2015neural,xu2015show}.
It aims to focus on certain aspects of data sequentially and aggregate essential information over time to infer the results, and has been successfully applied to both areas of vision and language.
In computer vision, attention based methods adaptively select a sequence of image regions to extract necessary features~\cite{mnih2014recurrent,gregor2015draw,xu2015show}.
Similarly, attention models for natural language processing highlight specific words or sentences to distill information from input text~\cite{bahdanau2015neural,rush2015neural,kumar2016ask}.
These approaches have improved the performance of wide applications in conjunction with deep architectures including convolutional neural networks (CNNs) and recurrent neural networks (RNNs).

\begin{figure}[t]
\begin{center}
\begin{subfigure}[b]{\linewidth}
\includegraphics[width=1.07\linewidth]{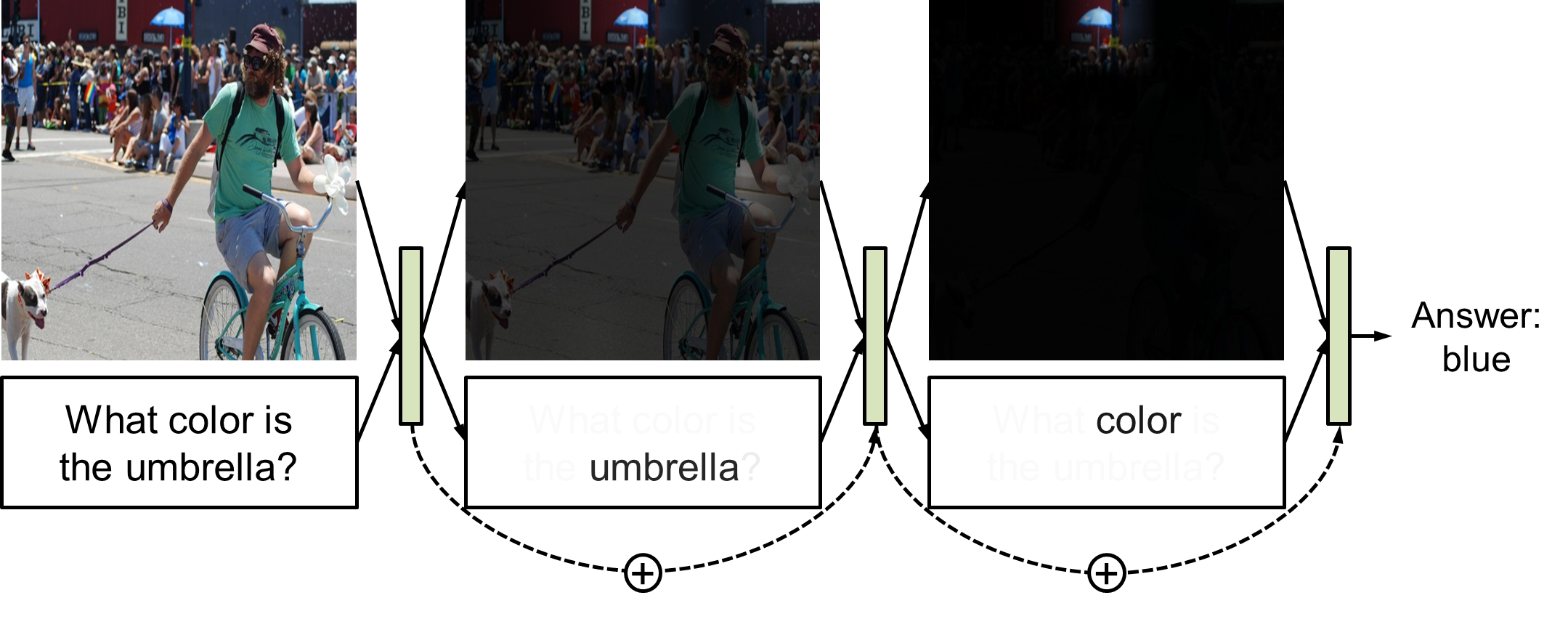}
\vspace{-5mm}
\caption{DAN for multimodal reasoning. (r-DAN)}
\label{fig:dan_vqa}
\vspace{3mm}
\end{subfigure}
\begin{subfigure}[b]{\linewidth}
\includegraphics[width=1.07\linewidth]{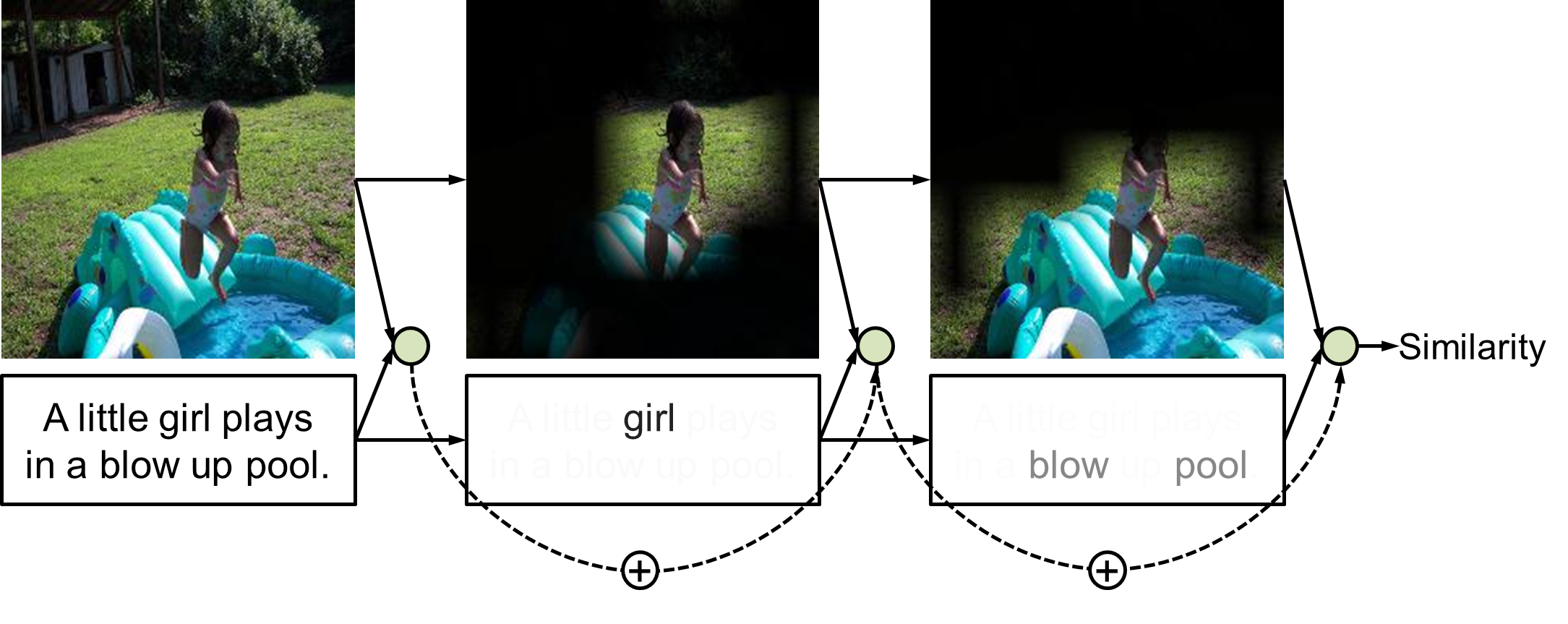}
\vspace{-5mm}
\caption{DAN for multimodal matching. (m-DAN)}
\label{fig:dan_matching}
\end{subfigure}
\caption{
Overview of Dual Attention Networks (DANs) for multimodal reasoning and matching.
The brightness of image regions and darkness of words indicate their attention weights predicted by DANs.
}
\label{fig:dan}
\vspace{-5mm}
\end{center}
\end{figure}

Despite the effectiveness of attention in handling both visual and textual data, it has been hardly attempted to establish a connection between visual and textual attention models which can be highly beneficial in various scenarios.
For example, the VQA problem in Figure~\ref{fig:dan_vqa} with the question {\tt What color is the umbrella?} can be efficiently solved by simultaneously focusing on the region of {\tt umbrella} and the word {\tt color}.
In the example of image-text matching in Figure~\ref{fig:dan_matching}, the similarity between the image and sentence can be effectively measured by attending to the specific regions and words sharing common semantics such as {\tt girl} and {\tt pool}.

In this paper, we propose \textit{Dual Attention Networks} (DANs) which jointly learn visual and textual attention models to explore the fine-grained interaction between vision and language.
We investigate two variants of DANs illustrated in Figure~\ref{fig:dan}, referred to as \textit{reasoning-DAN} (r-DAN) and \textit{matching-DAN} (m-DAN), respectively.
The r-DAN collaboratively performs visual and textual attentions using a joint memory which assembles the previous attention results and guides the next attentions.
It is suited to the tasks requiring multimodal reasoning such as VQA.
On the other hand, the m-DAN separates visual and textual attention models with distinct memories but jointly trains them to capture the shared semantics between images and sentences.
This approach eventually finds a joint embedding space which facilitates efficient cross-modal matching and retrieval.
Both proposed algorithms closely connect visual and textual attention mechanisms into a unified framework, achieving outstanding performance in VQA and image-text matching problems.

To summarize, the main contributions of our work are as follows:
\begin{itemize}
\item We propose an integrated framework of visual and textual attentions, where critical regions and words are jointly located through multiple steps.
\item Two variants of the proposed framework are implemented for multimodal reasoning and matching, and applied to VQA and image-text matching.
\item Detailed visualization of the attention results validates that our models effectively focus on vital portions of visual and textual data for the given task.
\item Our framework demonstrates the state-of-the-art performance on the VQA dataset~\cite{antol2015vqa} and the Flickr30K image-text matching dataset~\cite{young2014image}.  
\end{itemize}

\section{Related Work}
\subsection{Attention Mechanisms}
Attention mechanisms allow models to focus on necessary parts of visual or textual inputs at each step of a task.
Visual attention models selectively pay attention to small regions in an image to extract core features as well as reduce the amount of information to process.
A number of methods have recently adopted visual attention to benefit image classification~\cite{mnih2014recurrent,stollenga2014deep}, image generation~\cite{gregor2015draw}, image captioning~\cite{xu2015show}, visual question answering~\cite{yang2016stacked,shih2016where,xiong2016dynamic}, etc.
On the other hand, textual attention mechanisms generally aim to find semantic or syntactic input-output alignments under an encoder-decoder framework, which is especially effective in handling long-term dependency.
This approach has been successfully applied to various tasks including machine translation~\cite{bahdanau2015neural}, text generation~\cite{li2015hierarchical}, sentence summarization~\cite{rush2015neural}, and question answering~\cite{kumar2016ask,xiong2016dynamic}.

\subsection{Visual Question Answering (VQA)}
VQA is a task of answering a question in natural language regarding a given image, which requires multimodal reasoning over visual and textual data.
It has received a surge of interest since Antol \etal~\cite{antol2015vqa} presented a large-scale dataset with free-form and open-ended questions.
A simple baseline by Zhou \etal~\cite{zhou2015simple} predicts the answer from a concatenation of CNN image features and bag-of-word question features.
Several methods adaptively construct a deep architecture depending on the given question. 
For example, Noh \etal~\cite{noh2016image} impose a dynamic parameter layer on a CNN which is learned by the question, while Andreas \etal~\cite{andreas2016neural} utilize a compositional structure of the question to assemble a collection of neural modules.

One limitation of the above approaches is that they resort to a global image representation which contains noisy or unnecessary information.
To address this problem, Yang \etal~\cite{yang2016stacked} propose stacked attention networks which perform multi-step visual attention, and Shih \etal~\cite{shih2016where} use object proposals to identify regions relevant to the given question.
Recently, dynamic memory networks \cite{xiong2016dynamic} integrate an attention mechanism with a memory module, and multimodal compact bilinear pooling~\cite{fukui2016multimodal} is exploited to expressively combine multimodal features and predict attention over the image.
These methods commonly employ visual attention to find critical regions, but textual attention has been rarely incorporated into VQA.
Although HieCoAtt~\cite{lu2016hierarchical} applies both visual and textual attentions, it independently performs each step of co-attention without reasoning over previous co-attention outputs.
On the contrary, our method moves and refines both attentions via multiple reasoning steps based on the memory of previous attentions, which facilitates close interplay between visual and textual data.

\subsection{Image-Text Matching}
The core issue with image-text matching is measuring the semantic similarity between visual and textual inputs.
It is commonly addressed by learning a joint space where image and sentence feature vectors are directly comparable.
Hodosh \etal~\cite{hodosh2013framing} apply canonical correlation analysis (CCA) to find embeddings that maximize the correlation between images and sentences, which is further improved by incorporating deep neural networks \cite{klein2015associating,yan2015deep}.
A recent approach by Wang \etal~\cite{wang2016learning} includes structure-preserving constraints within a bidirectional loss function to make the joint space more discriminative.
In contrast, Ma \etal~\cite{ma2015multimodal} construct a CNN to combine an image and sentence fragments into a joint representation, from which the matching score is directly inferred.
Image captioning frameworks are also exploited to estimate the similarity based on the inverse probability of sentences given a query image \cite{mao2015deep,vinyals2015show}.

To the best of our knowledge, no study has attempted to learn multimodal attention models for image-text matching. 
Even though Karpathy \etal~\cite{karpathy2014deep,karpathy2015deep} propose to find the alignments between image regions and sentence fragments, they explicitly compute all pairwise distances between them and estimate the average or best alignment score, which leads to inefficiency.
On the other hand, our method automatically attends to the shared concepts between images and sentences while embedding them into a joint space, where cross-modal similarity is directly obtained by a single inner product operation.

\section{Dual Attention Networks (DANs)}
We present two structures of DANs to consolidate visual and textual attention mechanisms: r-DAN for multimodal reasoning and m-DAN for multimodal matching.
They share a common framework but differ in their ways of associating visual and textual attentions.
We first describe the common framework including input representation (Section~\ref{sub:input}) and attention mechanisms (Section~\ref{sub:attention}).
Then we illustrate the details of r-DAN (Section~\ref{sub:vqa}) and m-DAN (Section~\ref{sub:matching}) applied to VQA and image-text matching, respectively.


\subsection{Input Representation}
\label{sub:input}

\paragraph{Image representation}
The image features are extracted from 19-layer VGGNet~\cite{simonyan2014very} or 152-layer ResNet~\cite{he2016deep}.
We first rescale images to 448$\times$448 and feed them into the CNNs.
In order to obtain feature vectors for different regions, we take the last pooling layer of VGGNet (pool5) or the layer beneath the last pooling layer of ResNet (res5c).
Finally the input image is represented by $\{\mathbf{v}_1,\cdots,\mathbf{v}_N\}$, where $N$ is the number of image regions and $\mathbf{v}_n$ is a 512 (VGGNet) or 2048 (ResNet) dimensional feature vector corresponding to the $n$-th region.

\begin{figure}[t]
\begin{center}
\includegraphics[width=\linewidth]{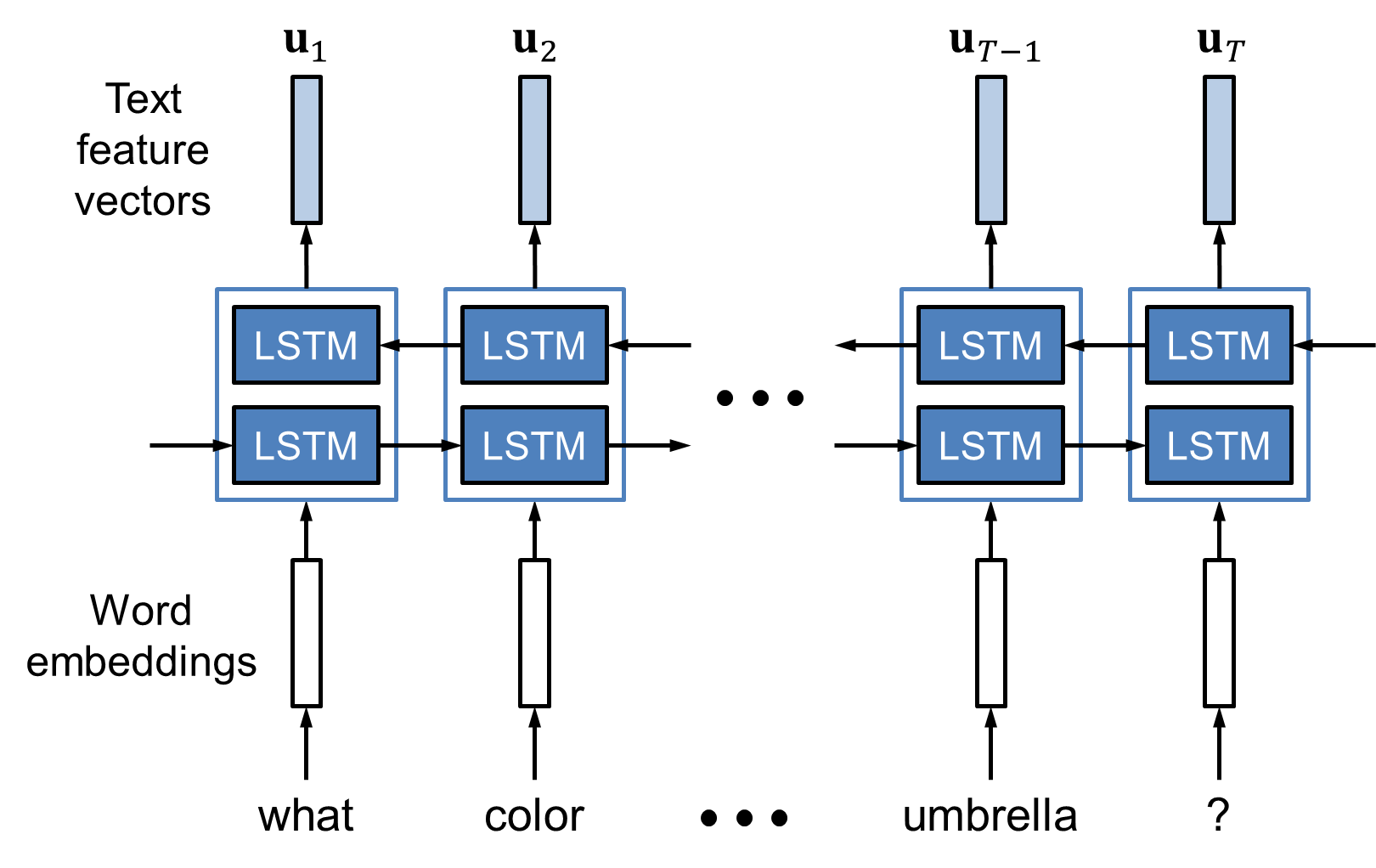}
\caption{
Bidirectional LSTMs for text encoding.
}
\label{fig:input_text}
\vspace{-5mm}
\end{center}
\end{figure}

\paragraph{Text representation}
We employ bidirectional LSTMs to generate text features as depicted in Figure~\ref{fig:input_text}.
Given one-hot encoding of $T$ input words $\{\mathbf{w}_1,\cdots,\mathbf{w}_T\}$, we first embed the words into a vector space by $\mathbf{x}_t = \mathbf{M}\mathbf{w}_t$, where $\mathbf{M}$ is an embedding matrix.
Then we feed the vectors into the bidirectional LSTMs:
\begin{align}
\label{eq:target}
\mathbf{h}_t^{(f)} &= \operatorname{\bf{LSTM}}^{(f)}(\mathbf{x}_t, \mathbf{h}_{t-1}^{(f)}), \\
\mathbf{h}_t^{(b)} &= \operatorname{\bf{LSTM}}^{(b)}(\mathbf{x}_t, \mathbf{h}_{t+1}^{(b)}),
\end{align}
where $\mathbf{h}_t^{(f)}$ and $\mathbf{h}_t^{(b)}$ represent the hidden states at time $t$ from the forward and backward LSTMs, respectively.
By adding the two hidden states at each time step, \ie $\mathbf{u}_t = \mathbf{h}_t^{(f)} + \mathbf{h}_t^{(b)}$, we construct a set of feature vectors $\{\mathbf{u}_1,\cdots,\mathbf{u}_T\}$ where $\mathbf{u}_t$ encodes the semantics of the $t$-th word in the context of the entire sentence.
Note that the models discussed here including the word embedding matrix and the LSTMs are trained end-to-end.

\subsection{Attention Mechanisms}
\label{sub:attention}
Our method performs visual and textual attentions simultaneously through multiple steps and gathers necessary information from both modalities.
In this section, we explain the underlying attention mechanisms employed at each step, which serve as building blocks to compose the entire DANs.
For simplicity, we shall omit the bias term $\mathbf{b}$ in the following equations.

\paragraph{Visual Attention.}
Visual attention aims to generate a context vector by attending to certain parts of the input image.
At step $k$, the visual context vector $\mathbf{v}^{(k)}$ is given by
\begin{align}
\mathbf{v}^{(k)} &= \operatorname{\bf{V\_Att}}(\{\mathbf{v}_n\}_{n=1}^N, \mathbf{m}_\mathbf{v}^{(k-1)}),
\label{eq:v_att}
\end{align}
where $\mathbf{m}_\mathbf{v}^{(k-1)}$ is a memory vector encoding the information that has been attended until step $k-1$.
Specifically, we employ the soft attention mechanism where the context vector is obtained from a weighted average of input feature vectors.
The attention weights $\{\alpha_{\mathbf{v},n}^{(k)}\}_{n=1}^N$ are computed by a 2-layer feed-forward neural network (FNN) and the softmax function:
\begin{align}
\mathbf{h}_{\mathbf{v},n}^{(k)} &= \tanh \left( \mathbf{W}_\mathbf{v}^{(k)}~\mathbf{v}_n \right) 
				   	     \odot \tanh \left( \mathbf{W}_{\mathbf{v},\mathbf{m}}^{(k)}~\mathbf{m}_\mathbf{v}^{(k-1)} \right), \\
\alpha_{\mathbf{v},n}^{(k)} &= \operatorname{softmax} \left( \mathbf{W}_{\mathbf{v},\mathbf{h}}^{(k)}~\mathbf{h}_{\mathbf{v},n}^{(k)} \right), \\
\mathbf{v}^{(k)} &= \tanh \left( \mathbf{P}^{(k)} \sum_{n=1}^N {\alpha_{\mathbf{v},n}^{(k)}~\mathbf{v}_n} \right),
\label{eq:v_context}
\end{align}
where $\mathbf{W}_\mathbf{v}^{(k)}$, $\mathbf{W}_{\mathbf{v},\mathbf{m}}^{(k)}$, and $\mathbf{W}_{\mathbf{v},\mathbf{h}}^{(k)}$ are the network parameters, $\mathbf{h}_{\mathbf{v},n}^{(k)}$ is a hidden state, and $\odot$ is element-wise multiplication.
In Equation~\ref{eq:v_context}, we introduce an additional layer with the weight matrix $\mathbf{P}^{(k)}$ in order to embed visual context vectors into a compatible space with textual context vectors, as we use pretrained image features $\mathbf{v}_n$.

\paragraph{Textual Attention.}
Textual attention computes a textual context vector $\mathbf{u}^{(k)}$ by focusing on specific words in the input sentence every step:
\begin{align}
\mathbf{u}^{(k)} &= \operatorname{\bf{T\_Att}}(\{\mathbf{u}_t\}_{t=1}^T, \mathbf{m}_\mathbf{u}^{(k-1)}),
\label{eq:t_att}
\end{align}
where $\mathbf{m}_\mathbf{u}^{(k-1)}$ is a memory vector.
The textual attention mechanism is almost identical to the visual attention mechanism.
In other words, the attention weights $\{\alpha_{\mathbf{u},t}^{(k)}\}_{t=1}^T$ are obtained from a 2-layer FNN and the context vector $\mathbf{u}^{(k)}$ is calculated by weighted averaging:
\begin{align}
\mathbf{h}_{\mathbf{u},t}^{(k)} &= \tanh \left( \mathbf{W}_\mathbf{u}^{(k)}~\mathbf{u}_t \right) 
				   	     \odot \tanh \left( \mathbf{W}_{\mathbf{u},\mathbf{m}}^{(k)}~\mathbf{m}_\mathbf{u}^{(k-1)} \right), \\
\alpha_{\mathbf{u},t}^{(k)} &= \operatorname{softmax} \left( \mathbf{W}_{\mathbf{u},\mathbf{h}}^{(k)}~\mathbf{h}_{\mathbf{u},t}^{(k)} \right), \\
\mathbf{u}^{(k)} &= \sum_{t}{\alpha_{\mathbf{u},t}^{(k)}~\mathbf{u}_t}.
\label{eq:t_context}
\end{align}
where $\mathbf{W}_\mathbf{u}^{(k)}$, $\mathbf{W}_{\mathbf{u},\mathbf{m}}^{(k)}$, and $\mathbf{W}_{\mathbf{u},\mathbf{h}}^{(k)}$ are the network parameters, $\mathbf{h}_{\mathbf{u},t}^{(k)}$ is a hidden state.
Unlike the visual attention, it does not need an additional layer after the last weighted averaging because the text features $\mathbf{u}_t$ are already trained end-to-end.

\subsection{r-DAN for Visual Question Answering}
\label{sub:vqa}
\begin{figure}[t]
\begin{center}
\includegraphics[width=1.05\linewidth]{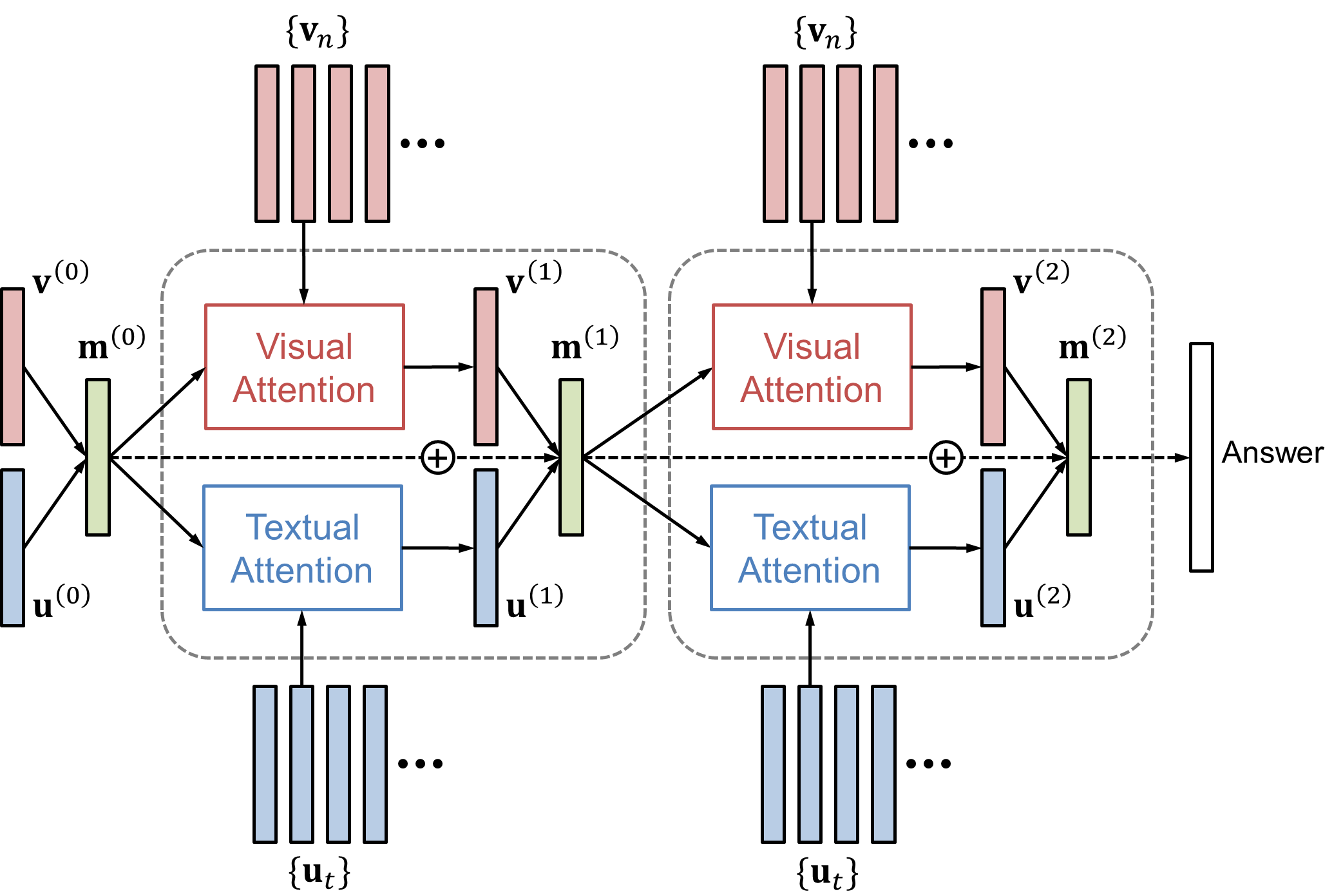}
\caption{
r-DAN in case of $K=2$.
}
\label{fig:dan_vqa_detail}
\vspace{-5mm}
\end{center}
\end{figure}

VQA is a representative problem which requires joint reasoning over multimodal data.
For this purpose, the r-DAN maintains a joint memory vector $\mathbf{m}^{(k)}$ which accumulates the visual and textual information that has been attended until step $k$.
It is recursively updated by 
\begin{align}
\mathbf{m}^{(k)} = \mathbf{m}^{(k-1)} + \mathbf{v}^{(k)} \odot \mathbf{u}^{(k)},
\label{eq:mem_vqa}
\end{align}
where $\mathbf{v}^{(k)}$ and $\mathbf{u}^{(k)}$ are the visual and textual context vectors obtained from Equation~\ref{eq:v_context} and \ref{eq:t_context}, respectively.
This joint representation concurrently guides the visual and textual attentions, \ie $\mathbf{m}^{(k)} = \mathbf{m}_\mathbf{v}^{(k)} = \mathbf{m}_\mathbf{u}^{(k)}$, which allows the two attention mechanisms to closely cooperate with each other.
The initial memory vector $\mathbf{m}^{(0)}$ is defined based on global context vectors $\mathbf{v}^{(0)}$ and $\mathbf{u}^{(0)}$ as
\begin{align}
\mathbf{m}^{(0)} &= \mathbf{v}^{(0)} \odot \mathbf{u}^{(0)}, \\
\text{where}~~
\mathbf{v}^{(0)} &= \tanh \left( \mathbf{P}^{(0)} \frac{1}{N} \sum_{n}{\mathbf{v}_n} \right), 
\label{eq:initial_v} \\
\mathbf{u}^{(0)} &= \frac{1}{T} \sum_{t}{\mathbf{u}_t}.
\label{eq:initial_u}
\end{align}
By repeating the dual attention (Equation~\ref{eq:v_att} and \ref{eq:t_att}) and memory update (Equation~\ref{eq:mem_vqa}) for $K$ steps, we effectively focus on the key portions  in the image and question, and gather relevant information for answering the question.
Figure~\ref{fig:dan_vqa_detail} illustrates the overall architecture of r-DAN in case of $K=2$.

The final answer is predicted by multi-way classification to the top $C$ frequent answers.
We employ a single-layer softmax classifier with cross-entropy loss where the input is the final memory $\mathbf{m}^{(K)}$:
\begin{align}
\mathbf{p}_{ans} = \operatorname{softmax} \left( \mathbf{W}_{ans}~\mathbf{m}^{(K)} \right),
\end{align}
where $\mathbf{p}_{ans}$ represents the probability over the candidate answers.

\subsection{m-DAN for Image-Text Matching}
\label{sub:matching}
Image-text matching tasks usually involve comparison between numerous images and sentences, where effective and efficient computation of cross-modal similarities is crucial.
To achieve this, we aim to learn a joint embedding space which satisfies the following properties.
First, the embedding space encodes the shared concepts that frequently co-occur in image and sentence domains.
Moreover, images and sentences are autonomously embedded into the joint space without being paired, so that arbitrary image and sentence vectors in the  space are directly comparable.



\begin{figure}[t]
\begin{center}
\includegraphics[width=1.05\linewidth]{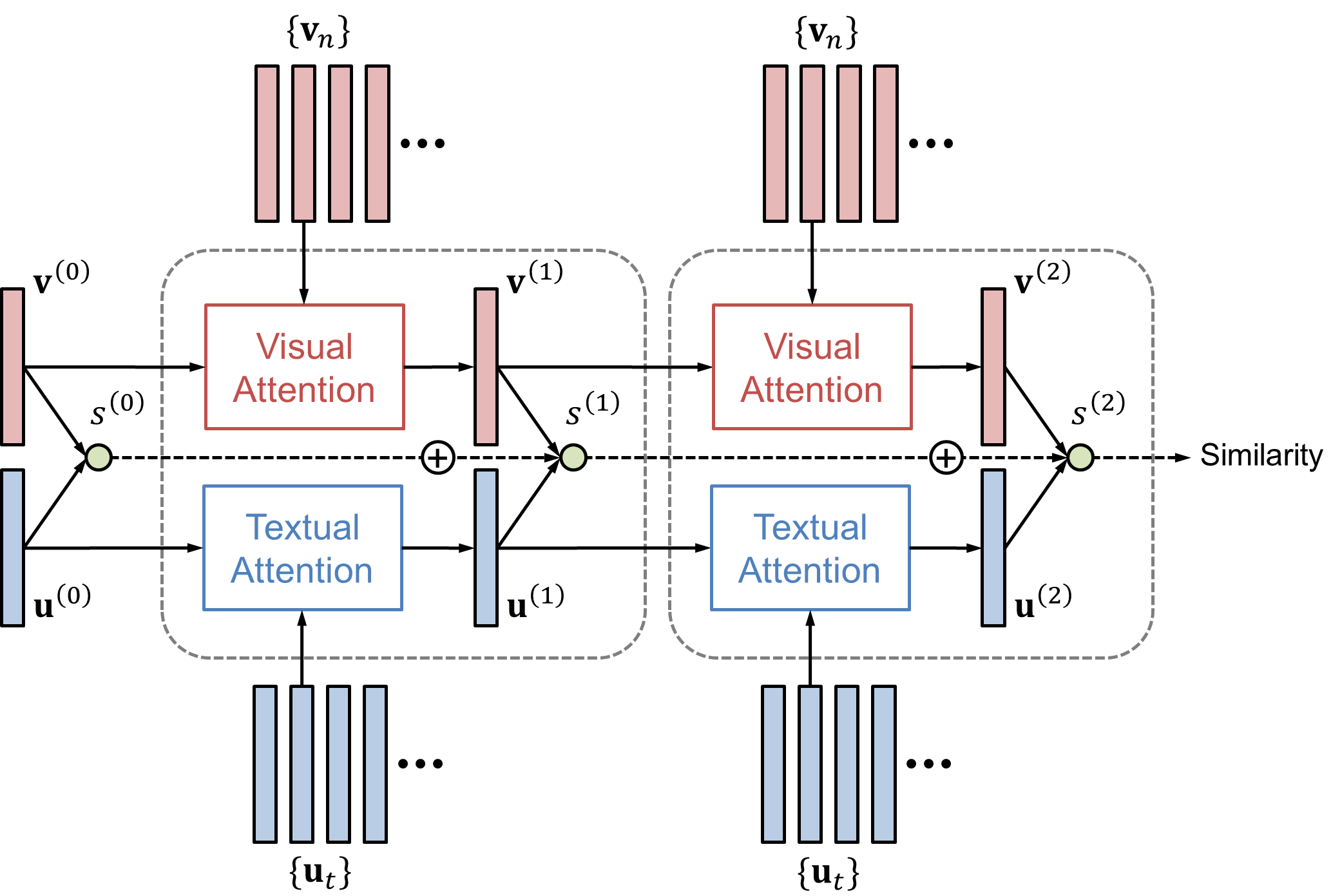}
\caption{
m-DAN in case of $K=2$.
}
\label{fig:dan_matching_detail}
\vspace{-5mm}
\end{center}
\end{figure}

Our m-DAN jointly learns visual and textual attention models to capture the shared concepts between the two modalities, but separates them at inference time to provide generally comparable representations in the embedding space.
Contrary to the r-DAN which uses a joint memory, the m-DAN maintains separate memory vectors for visual and textual attentions as follows:
\begin{align}
\mathbf{m}_\mathbf{v}^{(k)} = \mathbf{m}_\mathbf{v}^{(k-1)} + \mathbf{v}^{(k)}, \\
\mathbf{m}_\mathbf{u}^{(k)} = \mathbf{m}_\mathbf{u}^{(k-1)} + \mathbf{u}^{(k)},
\label{eq:mem_matching}
\end{align}
which are initialized to $\mathbf{v}^{(0)}$ and $\mathbf{u}^{(0)}$ defined in Equation~\ref{eq:initial_v} and \ref{eq:initial_u}, respectively.
At each step, we compute the similarity $s^{(k)}$ between visual and textual context vectors by their inner product:
\begin{align}
s^{(k)} = \mathbf{v}^{(k)} \cdot \mathbf{u}^{(k)}.
\end{align}
After performing $K$ steps of the dual attention and memory update, the final similarity $S$ between the given image and sentence becomes
\begin{align}
S = \sum_{k=0}^{K} s^{(k)}.
\label{eq:sim}
\end{align}
The overall architecture of this model when $K=2$ is depicted in Figure~\ref{fig:dan_matching_detail}.

This network is trained with bidirectional max-margin ranking loss, which is widely adopted for multimodal similarity learning~\cite{karpathy2014deep,karpathy2015deep,kiros2015unifying,wang2016learning}.
For each correct pair of an image and a sentence $(\mathbf{v},\mathbf{u})$, we additionally sample a negative image $\mathbf{v}^-$ and a negative sentence $\mathbf{u}^-$ to construct two negative pairs $(\mathbf{v}^-,\mathbf{u})$ and $(\mathbf{v},\mathbf{u}^-)$.
Then, the loss function becomes:
\begin{align}
\mathcal{L} = \sum_{(\mathbf{v},\mathbf{u})} \Big\{ 
		     \max \left[ 0, m - S(\mathbf{v},\mathbf{u}) + S(\mathbf{v}^-,\mathbf{u}) \right] \nonumber \\
		    + \max \left[ 0, m - S(\mathbf{v},\mathbf{u}) + S(\mathbf{v},\mathbf{u}^-) \right] \Big\},
\label{eq:loss_matching}
\end{align}
where $m$ is a margin constraint.
By minimizing this function, the network is trained to focus on the common semantics that only appears in correct image-sentence pairs through visual and textual attention mechanisms.

At inference time, an arbitrary image or sentence is embedded into the joint space by concatenating its context vectors:
\begin{align}
	\mathbf{z}_\mathbf{v} &= [\mathbf{v}^{(0)};\cdots;\mathbf{v}^{(K)}], \\
	\mathbf{z}_\mathbf{u} &= [\mathbf{u}^{(0)};\cdots;\mathbf{u}^{(K)}],
\end{align}
where $\mathbf{z}_\mathbf{v}$ and $\mathbf{z}_\mathbf{u}$ are the representations for image $\mathbf{v}$ and sentence $\mathbf{u}$, respectively.
Note that these vectors are obtained via separate pipelines of visual and textual attentions, \ie learned shared concepts are revealed from an image or sentence itself, not from an image-sentence pair.
The similarity between two vectors in the joint space is simply computed by their inner product, \eg $S(\mathbf{v}, \mathbf{u}) = \mathbf{z}_\mathbf{v} \cdot \mathbf{z}_\mathbf{u}$, which is equivalent to the output of the network in Equation~\ref{eq:sim}.

\section{Experiments}

\subsection{Experimental Setup}
We fix all the hyper-parameters applied to both r-DAN and m-DAN.
The number of attention steps $K$ is set to 2 which empirically shows the best performance.
The dimension of every hidden layer---including word embedding, LSTMs, and attention models---is set to 512.
We train our networks by stochastic gradient descent with a learning rate 0.1, momentum 0.9, weight decay 0.0005, dropout ratio 0.5, and gradient clipping at 0.1.
The network is trained for 60 epochs, where the learning rate is dropped to 0.01 after 30 epochs.
A minibatch for r-DAN and m-DAN consists of 128 pairs of $\langle$image, question$\rangle$ and 128 quadruplets of $\langle$positive image, positive sentence, negative image, negative sentence$\rangle$, respectively.
The number of possible answers $C$ for VQA is set to 2000, and the margin $m$ for the loss function in Equation~\ref{eq:loss_matching} is set to 100.

\begin{table*}[t]
\caption{
Results on the VQA dataset compared with state-of-the-art methods.
}
\vspace{-2mm}
  \centering
  \begin{tabular}{lcccccccccc}
    \toprule
	& \multicolumn{5}{c}{Test-dev}
	& \multicolumn{5}{c}{Test-standard} \\
    \cmidrule(lr){2-6} \cmidrule(lr){7-11}
	& \multicolumn{4}{c}{Open-Ended}
	& {MC} 
	& \multicolumn{4}{c}{Open-Ended}
	& {MC} \\
    \cmidrule(lr){2-5} \cmidrule(lr){6-6} \cmidrule(lr){7-10} \cmidrule(lr){11-11}
	{Method}
	& {Y/N} & {Num} & {Other} & {All} & {All}
	& {Y/N} & {Num} & {Other} & {All} & {All} \\
    \midrule
	{iBOWIMG~\cite{zhou2015simple}}
	& {76.5} & {35.0} & {42.6} & {55.7} & {61.7}
	& {76.8} & {35.0} & {42.6} & {55.9} & {62.0} \\
	{DPPnet~\cite{noh2016image}}
	& {80.7} & {37.2} & {41.7} & {57.2} & {62.5}
	& {80.3} & {36.9} & {42.2} & {57.4} & {62.7} \\
	{VQA team~\cite{antol2015vqa}}
	& {80.5} & {36.8} & {43.1} & {57.8} & {62.7}
	& {80.6} & {36.5} & {43.7} & {58.2} & {63.1} \\
	{SAN~\cite{yang2016stacked}}
	& {79.3} & {36.6} & {46.1} & {58.7} & {-}
	& {-} & {-} & {-} & {58.9} & {-} \\
	{NMN~\cite{andreas2016neural}}
	& {81.2} & {38.0} & {44.0} & {58.6} & {-}
	& {-} & {-} & {-} & {58.7} & {-} \\
	{ACK~\cite{wu2016ask}}
	& {81.0} & {38.4} & {45.2} & {59.2} & {-}
	& {81.1} & {37.1} & {45.8} & {59.4} & {-} \\
	{DMN+~\cite{xiong2016dynamic}}
	& {80.5} & {36.8} & {48.3} & {60.3} & {-}
	& {-} & {-} & {-} & {60.4} & {-} \\
	{MRN (ResNet)~\cite{kim2016multimodal}}
	& {82.3} & {38.8} & {49.3} & {61.7} & {66.2}
	& {82.4} & {\bf{38.2}} & {49.4} & {61.8} & {66.3} \\
	{HieCoAtt (ResNet)~\cite{lu2016hierarchical}}
	& {79.7} & {38.7} & {51.7} & {61.8} & {65.8}
	& {-} & {-} & {-} & {62.1} & {66.1} \\
	{RAU (ResNet)~\cite{noh2016training}}
	& {81.9} & {39.0} & {53.0} & {63.3} & {67.7}
	& {81.7} & {\bf{38.2}} & {52.8} & {63.2} & {67.3} \\
	{MCB (ResNet)~\cite{fukui2016multimodal}}
	& {82.2} & {37.7} & {\bf{54.8}} & {64.2} & {68.6}
	& {-} & {-} & {-} & {-} & {-} \\
	\midrule
	{DAN\ (VGG)}
	& {82.1} & {38.2} & {50.2} & {62.0} & {67.0}
	& {-} & {-} & {-} & {-} & {-} \\
	{DAN\ (ResNet)}
	& {\bf{83.0}} & {\bf{39.1}} & {53.9} & {\bf{64.3}} & {\bf{69.1}}
	& {\bf{82.8}} & {38.1} & {\bf{54.0}} & {\bf{64.2}} & {\bf{69.0}} \\
    \hline
  \end{tabular}
\label{tab:vqa}
\end{table*}

\begin{table*}
\centering
\setlength\tabcolsep{0pt}

\begin{tabular}{P{0.166\linewidth}P{0.166\linewidth}P{0.166\linewidth}P{0.01\linewidth}P{0.166\linewidth}P{0.166\linewidth}P{0.166\linewidth}}
\includegraphics[width=\linewidth, height=25mm]{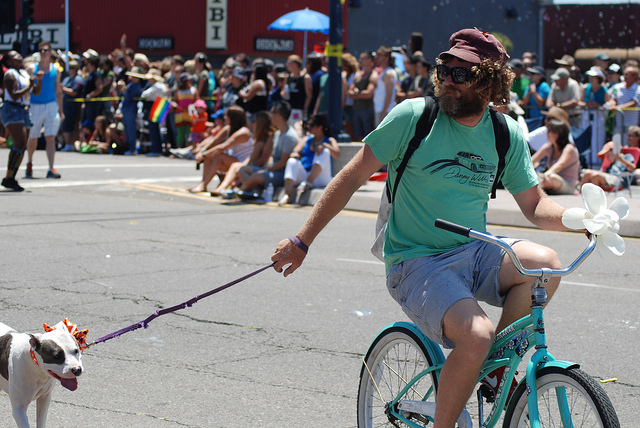} &
\includegraphics[width=\linewidth, height=25mm]{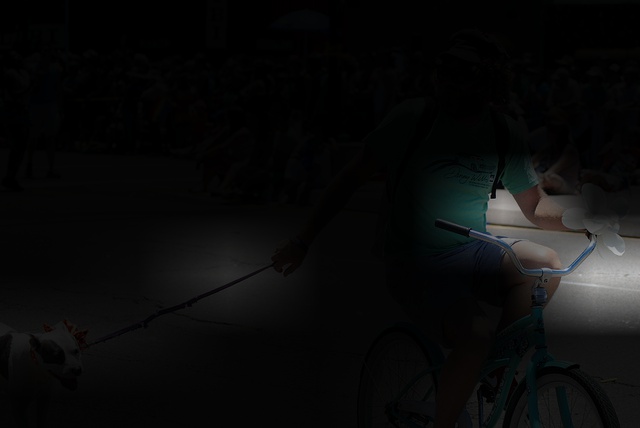} &
\includegraphics[width=\linewidth, height=25mm]{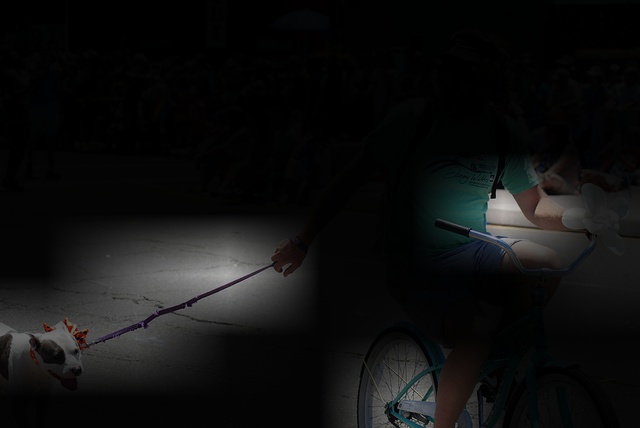} &
&
\includegraphics[width=\linewidth, height=25mm]{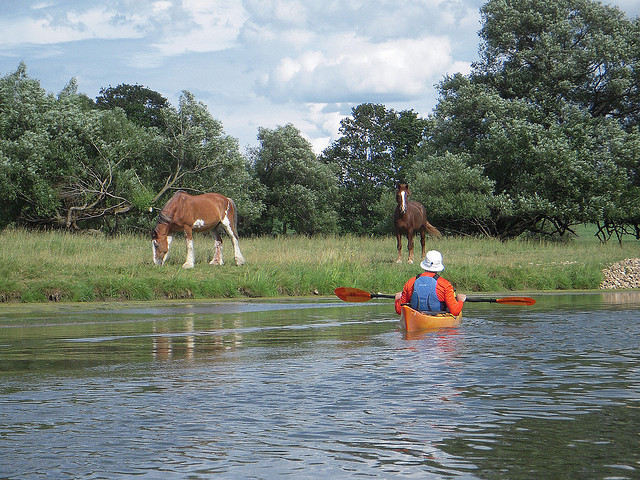} &
\includegraphics[width=\linewidth, height=25mm]{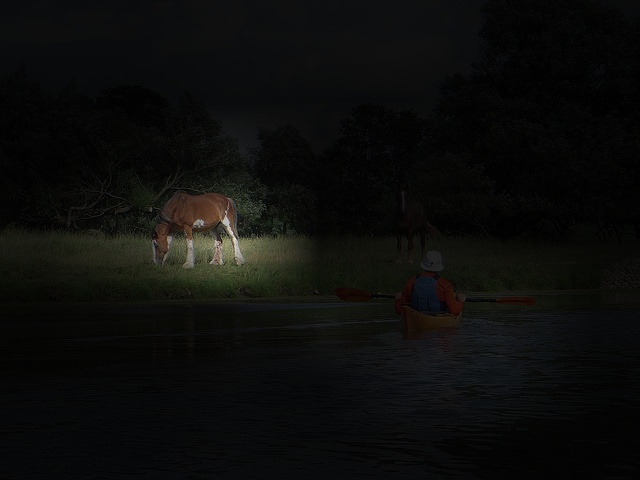} &
\includegraphics[width=\linewidth, height=25mm]{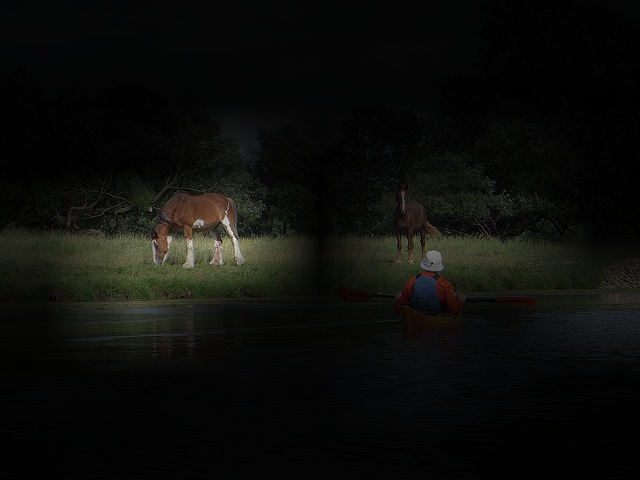}
\\
{\textbf{\g[0]Q:} What is the man on the bike holding on his right hand? \newline \textbf{A:} leash} \vspace{1mm} &
{\textbf{\g[1]Q:} \g[0.99]What \g[0.99]is \g[0.99]the \g[0.99]man \g[0.99]on \g[0.99]the \g[0.99]bike \g[0.60]holding \g[0.99]on \g[0.99]his \g[0.99]right \g[0.51]hand\g[0.99]?}&
{\textbf{\g[1]Q:} \g[0.99]What \g[0.99]is \g[0.99]the \g[0.99]man \g[0.99]on \g[0.99]the \g[0.98]bike \g[0.58]holding \g[0.99]on \g[0.99]his \g[0.99]right \g[0.55]hand\g[0.99]?}&
&
{\textbf{\g[0]Q:} How many horses are in the picture? \newline \textbf{A:} 2}&
{\textbf{\g[1]Q:} \g[0.93]How \g[0.95]many \g[0.33]horses \g[0.95]are \g[0.95]in \g[0.95]the \g[0.95]picture\g[0.95]?}&
{\textbf{\g[1]Q:} \g[0.71]How \g[0.51]many \g[0.96]horses \g[0.97]are \g[0.96]in \g[0.97]the \g[0.96]picture\g[0.97]?}
\\

\includegraphics[width=\linewidth, height=25mm]{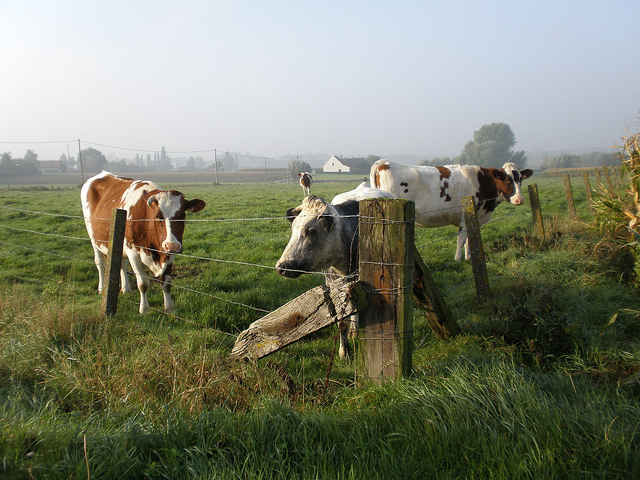} &
\includegraphics[width=\linewidth, height=25mm]{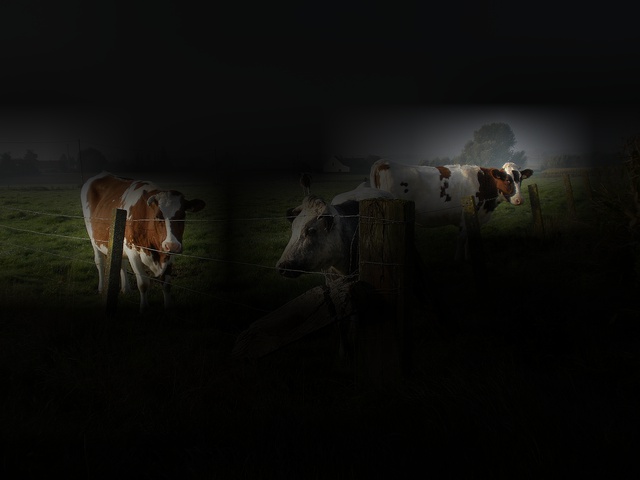} &
\includegraphics[width=\linewidth, height=25mm]{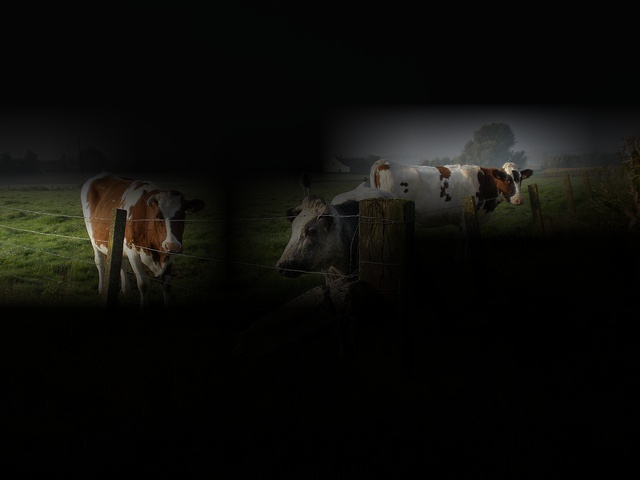} &
&
\includegraphics[width=\linewidth, height=25mm]{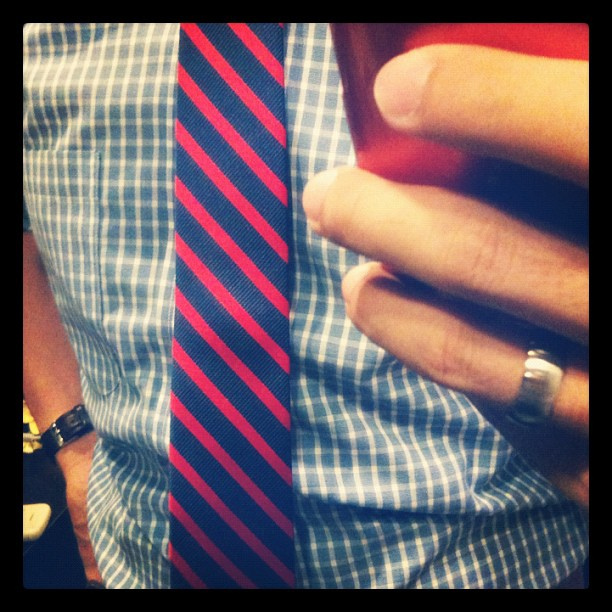} &
\includegraphics[width=\linewidth, height=25mm]{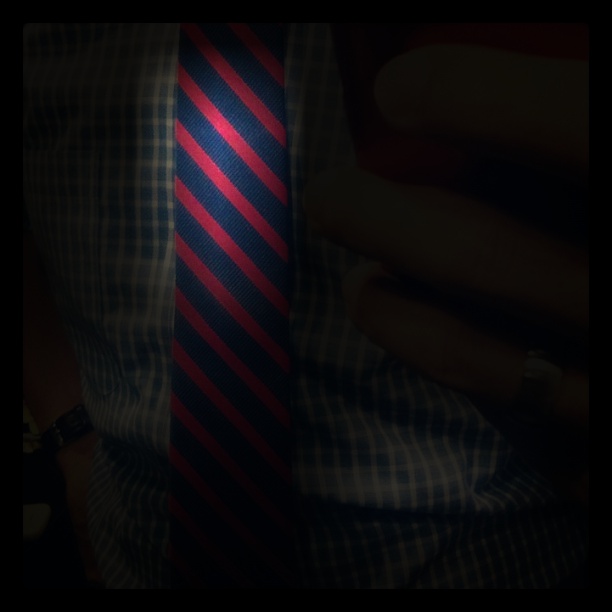} &
\includegraphics[width=\linewidth, height=25mm]{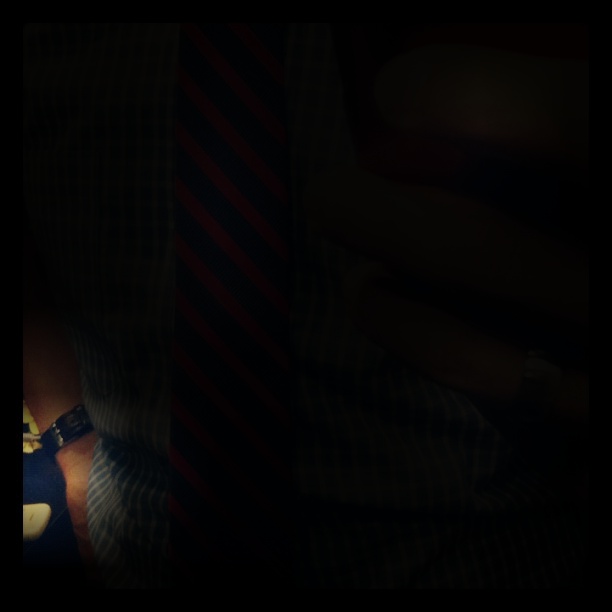}
\\
{\textbf{\g[0]Q:} What color are the cows?\newline \textbf{A:} brown and white}&
{\textbf{\g[1]Q:} \g[0.93]What \g[0.93]color \g[0.93]are \g[0.93]the \g[0.39]cows\g[0.88]?}&
{\textbf{\g[1]Q:} \g[0.95]What \g[0.26]color \g[0.95]are \g[0.95]the \g[0.94]cows\g[0.95]?} &
&
{\textbf{\g[0]Q:} What is on his wrist? \newline \textbf{A:} watch} \vspace{1mm} &
{\textbf{\g[1]Q:} \g[0.93]What \g[0.93]is \g[0.93]on \g[0.93]his \g[0.08]wrist\g[0.93]?}&
{\textbf{\g[1]Q:} \g[0.93]What \g[0.93]is \g[0.93]on \g[0.93]his \g[0.09]wrist\g[0.93]?}
\\
\end{tabular}
\vspace{-2mm}
\captionof{figure}{
Qualitative results on the VQA dataset with attention visualization.
For each example, the query image, question, and the answer by DAN are presented from top to bottom; the original image (question), the first and second attention maps are shown from left to right.
The brightness of images and darkness of words represent their attention weights.
}
\label{fig:vqa_qual}
\end{table*}

\begin{table*}
\caption{
Bidirectional retrieval results on the Flickr30K dataset compared with state-of-the-art methods.
}
\vspace{-2mm}
  \centering
  \begin{tabular}{lcccccccc}
    \toprule
	& \multicolumn{4}{c}{Image-to-Text}
	& \multicolumn{4}{c}{Text-to-Image} \\
    \cmidrule(lr){2-5} \cmidrule(lr){6-9}
	{Method}
	& {R@1} & {R@5} & {R@10} & {MR}
	& {R@1} & {R@5} & {R@10} & {MR}  \\
    \midrule
	{DCCA~\cite{yan2015deep}}
	& {27.9} & {56.9} & {68.2} & {4}
	& {26.8} & {52.9} & {66.9} & {4} \\
	{mCNN~\cite{ma2015multimodal}}
	& {33.6} & {64.1} & {74.9} & {3}
	& {26.2} & {56.3} & {69.6} & {4} \\
	{m-RNN-VGG~\cite{mao2015deep}}
	& {35.4} & {63.8} & {73.7} & {3}
	& {22.8} & {50.7} & {63.1} & {5} \\
	{GMM+HGLMM FV~\cite{klein2015associating}}
	& {35.0} & {62.0} & {73.8} & {3}
	& {25.0} & {52.7} & {66.0} & {5} \\
	{HGLMM FV~\cite{plummer2015flickr30k}}
	& {36.5} & {62.2} & {73.3} & {-}
	& {24.7} & {53.4} & {66.8} & {-} \\
	{SPE~\cite{wang2016learning}}
	& {40.3} & {68.9} & {79.9} & {-}
	& {29.7} & {60.1} & {72.1} & {-} \\
    \midrule
	{DAN\ (VGG)}
	& {41.4} & {73.5} & {82.5} & {2}
	& {31.8} & {61.7} & {72.5} & {3} \\
	{DAN\ (ResNet)}
	& {\bf{55.0}} & {\bf{81.8}} & {\bf{89.0}} & {\bf{1}}
	& {\bf{39.4}} & {\bf{69.2}} & {\bf{79.1}} & {\bf{2}} \\
    \hline
  \end{tabular}
\label{tab:matching}
\vspace{-3mm}
\end{table*}

\begin{table*}
\centering
\setlength\tabcolsep{0pt}

\begin{tabular}{C{0.166\linewidth}C{0.166\linewidth}C{0.166\linewidth}C{0.01\linewidth}C{0.166\linewidth}C{0.166\linewidth}C{0.166\linewidth}}
\includegraphics[width=\linewidth, height=25mm]{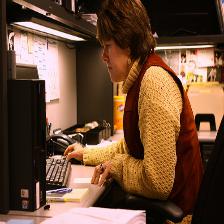} &
\includegraphics[width=\linewidth, height=25mm]{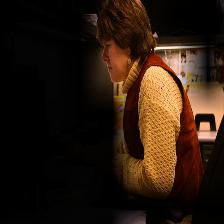} &
\includegraphics[width=\linewidth, height=25mm]{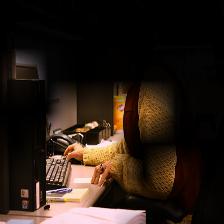} &
&
\includegraphics[width=\linewidth, height=25mm]{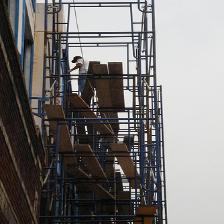} &
\includegraphics[width=\linewidth, height=25mm]{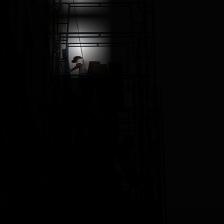} &
\includegraphics[width=\linewidth, height=25mm]{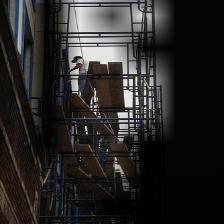}
\\
\hline
{\textbf{(+)} A woman in a brown vest is working on the computer.} &
{\g[1]\textbf{(+)} \g[0.94]A \g[0.57]woman \g[0.94]in \g[0.94]a \g[0.94]brown \g[0.94]vest \g[0.94]is \g[0.94]working \g[0.94]on \g[0.94]the \g[0.94]computer\g[1].}&
{\g[1]\textbf{(+)} \g[0.98]A \g[0.98]woman \g[0.98]in \g[0.98]a \g[0.98]brown \g[0.98]vest \g[0.98]is \g[0.98]working \g[0.97]on \g[0.98]the \g[0.16]computer\g[1].}&
&
{\textbf{(+)} A man in a white shirt stands high up on scaffolding.} &
{\g[1]\textbf{(+)} \g[0.99]A \g[0.68]man \g[0.99]in \g[0.99]a \g[0.68]white \g[0.99]shirt \g[0.99]stands \g[0.99]high \g[0.99]up \g[0.99]on \g[0.68]scaffolding\g[1].}&
{\g[1]\textbf{(+)} \g[0.99]A \g[0.99]man \g[0.99]in \g[0.99]a \g[0.99]white \g[0.99]shirt \g[0.99]stands \g[0.99]high \g[0.99]up \g[0.99]on \g[0.15]scaffolding\g[1].}
\\
\hline
{\textbf{(+)} A woman in a red vest working at a computer.} &
{\g[1]\textbf{(+)} \g[0.99]A \g[0.54]woman \g[0.99]in \g[0.99]a \g[0.99]red \g[0.54]vest \g[0.99]working \g[0.99]at \g[0.99]a \g[0.99]computer\g[1].}&
{\g[1]\textbf{(+)} \g[0.99]A \g[0.99]woman \g[0.99]in \g[0.99]a \g[0.99]red \g[0.99]vest \g[0.53]working \g[0.99]at \g[0.99]a \g[0.53]computer\g[1].}&
&
{\textbf{(+)} Man works on top of scaffolding.} &
{\g[1]\textbf{(+)} \g[0.52]Man \g[0.99]works \g[0.99]on \g[0.99]top \g[0.99]of \g[0.52]scaffolding\g[1].}&
{\g[1]\textbf{(+)} \g[0.98]Man \g[0.98]works \g[0.98]on \g[0.98]top \g[0.98]of \g[0.08]scaffolding\g[1].}
\\
\hline

\includegraphics[width=\linewidth, height=25mm]{matching_img2txt_0987_img.jpg} &
\includegraphics[width=\linewidth, height=25mm]{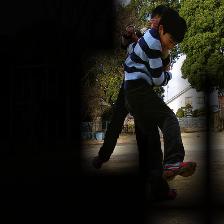} &
\includegraphics[width=\linewidth, height=25mm]{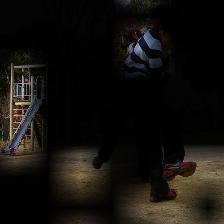} &
&
\includegraphics[width=\linewidth, height=25mm]{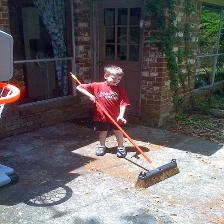} &
\includegraphics[width=\linewidth, height=25mm]{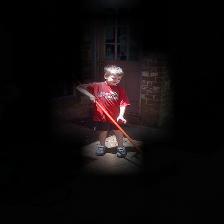} &
\includegraphics[width=\linewidth, height=25mm]{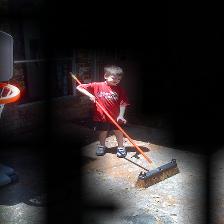}
\\
\hline
{\textbf{(+)} Two boys playing together at a playground.}&
{\g[1]\textbf{(+)} \g[0.99]Two \g[0.41]boys \g[0.99]playing \g[0.99]together \g[0.99]at \g[0.99]a \g[0.64]playground\g[1].}&
{\g[1]\textbf{(+)} \g[0.98]Two \g[0.98]boys \g[0.98]playing \g[0.98]together \g[0.98]at \g[0.98]a \g[0.01]playground\g[1].} &
&
{\textbf{(-)} A man wearing a red t shirt sweeps the sidewalk in front of a brick building.}&
{\g[1]\textbf{(-)} \g[0.99]A \g[0.69]man \g[0.99]wearing \g[0.99]a \g[0.69]red \g[0.99]t \g[0.99]shirt \g[0.69]sweeps \g[0.99]the \g[0.99]sidewalk \g[0.99]in \g[0.99]front \g[0.99]of \g[0.99]a \g[0.99]brick \g[0.99]building\g[1].}&
{\g[1]\textbf{(-)} \g[0.99]A \g[0.98]man \g[0.99]wearing \g[0.99]a \g[0.99]red \g[0.99]t \g[0.99]shirt \g[0.22]sweeps \g[0.99]the \g[0.99]sidewalk \g[0.99]in \g[0.99]front \g[0.99]of \g[0.99]a \g[0.99]brick \g[0.99]building\g[1].}
\\
\hline
{\textbf{(-)} The two kids are playing at the playground.}&
{\g[1]\textbf{(-)} \g[0.99]The \g[0.99]two \g[0.52]kids \g[0.99]are \g[0.99]playing \g[0.99]at \g[0.99]the \g[0.53]playground\g[1].}&
{\g[1]\textbf{(-)} \g[0.98]The \g[0.98]two \g[0.98]kids \g[0.98]are \g[0.98]playing \g[0.98]at \g[0.98]the \g[0.12]playground\g[1].} &
&
{\textbf{(+)} Boy in red shirt and black shorts sweeps driveway.}&
{\g[1]\textbf{(+)} \g[0.68]Boy \g[0.99]in \g[0.68]red \g[0.99]shirt \g[0.99]and \g[0.99]black \g[0.99]shorts \g[0.68]sweeps \g[0.99]driveway\g[1].}&
{\g[1]\textbf{(+)} \g[0.98]Boy \g[0.98]in \g[0.98]red \g[0.98]shirt \g[0.98]and \g[0.98]black \g[0.98]shorts \g[0.15]sweeps \g[0.95]driveway\g[1].}
\\
\hline
\end{tabular}
\vspace{-2mm}
\captionof{figure}{
Qualitative results from image-to-text retrieval with attention visualization.
For each example, the query image and the top two retrieved sentences are shown from top to bottom; the original image (sentence), the first and second attention maps are shown from left to right.
(+) and (-) indicate ground-truth and non ground-truth sentences, respectively.
}
\label{fig:img2txt}
\end{table*}

\begin{table*}
\centering
\setlength\tabcolsep{0pt}
\renewcommand{\arraystretch}{0}
\begin{tabular}{C{0.166\linewidth}C{0.166\linewidth}C{0.166\linewidth}C{0.01\linewidth}C{0.166\linewidth}C{0.166\linewidth}C{0.166\linewidth}}
\hline
\\
{A woman in a cap at a coffee shop.} &
{\g[0.99]A \g[0.69]woman \g[0.99]in \g[0.99]a \g[0.67]cap \g[0.99]at \g[0.99]a \g[0.68]coffee \g[0.99]shop\g[1].}&
{\g[0.98]A \g[0.98]woman \g[0.98]in \g[0.98]a \g[0.98]cap \g[0.98]at \g[0.98]a \g[0.13]coffee \g[0.98]shop\g[1].}&
&
{A boy is hanging out of the window of a yellow taxi.} &
{\g[0.98]A \g[0.17]boy \g[0.98]is \g[0.98]hanging \g[0.98]out \g[0.98]of \g[0.98]the \g[0.98]window \g[0.98]of \g[0.98]a \g[0.98]yellow \g[0.98]taxi\g[1].}&
{\g[0.99]A \g[0.99]boy \g[0.99]is \g[0.99]hanging \g[0.99]out \g[0.99]of \g[0.99]the \g[0.99]window \g[0.99]of \g[0.99]a \g[0.56]yellow \g[0.52]taxi\g[1].}
\\
\hline
\\
\includegraphics[width=\linewidth, height=25mm]{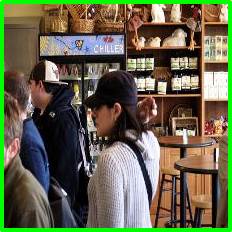} &
\includegraphics[width=\linewidth, height=25mm]{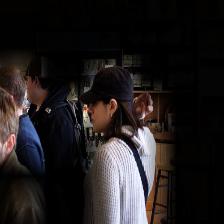} &
\includegraphics[width=\linewidth, height=25mm]{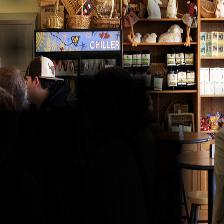} &
&
\includegraphics[width=\linewidth, height=25mm]{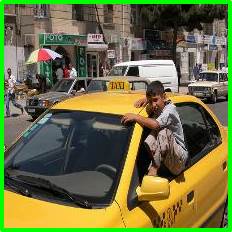} &
\includegraphics[width=\linewidth, height=25mm]{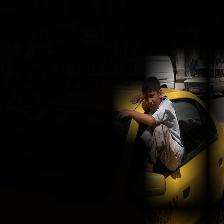} &
\includegraphics[width=\linewidth, height=25mm]{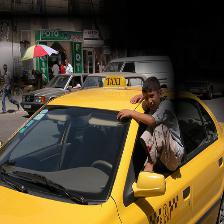}
\\
\includegraphics[width=\linewidth, height=25mm]{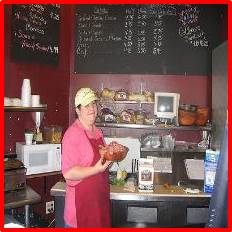} &
\includegraphics[width=\linewidth, height=25mm]{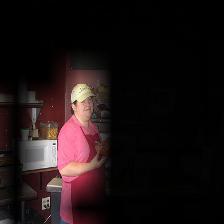} &
\includegraphics[width=\linewidth, height=25mm]{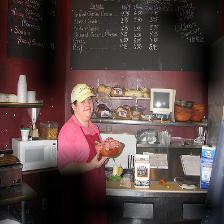} &
&
\includegraphics[width=\linewidth, height=25mm]{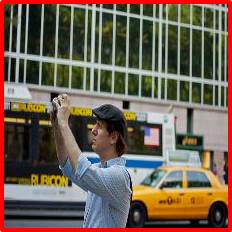} &
\includegraphics[width=\linewidth, height=25mm]{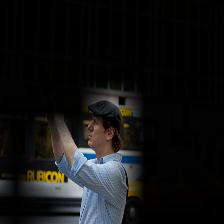} &
\includegraphics[width=\linewidth, height=25mm]{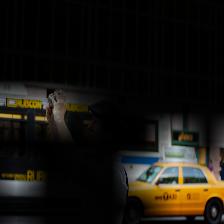}
\\\\\\
\hline
\\
{A woman in a striped outfit on a bike.} &
{\g[0.98]A \g[0.14]woman \g[0.98]in \g[0.98]a \g[0.98]striped \g[0.98]outfit \g[0.98]on \g[0.98]a \g[0.98]bike\g[1].}&
{\g[0.98]A \g[0.98]woman \g[0.98]in \g[0.98]a \g[0.98]striped \g[0.98]outfit \g[0.98]on \g[0.98]a \g[0.13]bike\g[1].}&
&
{A group of people standing on a sidewalk under some trees.} &
{\g[0.99]A \g[0.54]group \g[0.99]of \g[0.54]people \g[0.99]standing \g[0.99]on \g[0.99]a \g[0.99]sidewalk \g[0.99]under \g[0.99]some \g[0.98]trees\g[1].}&
{\g[0.97]A \g[0.97]group \g[0.97]of \g[0.89]people \g[0.97]standing \g[0.97]on \g[0.97]a \g[0.39]sidewalk \g[0.97]under \g[0.97]some \g[0.97]trees\g[1].}
\\\\
\hline
\\
\includegraphics[width=\linewidth, height=25mm]{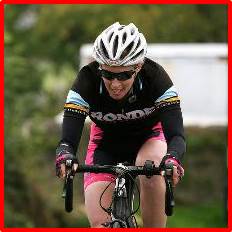} &
\includegraphics[width=\linewidth, height=25mm]{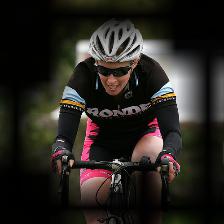} &
\includegraphics[width=\linewidth, height=25mm]{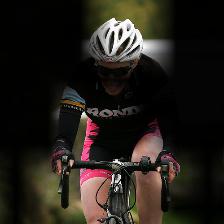} &
&
\includegraphics[width=\linewidth, height=25mm]{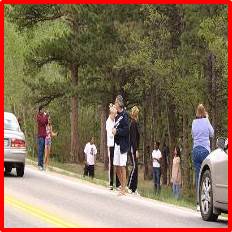} &
\includegraphics[width=\linewidth, height=25mm]{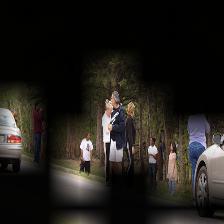} &
\includegraphics[width=\linewidth, height=25mm]{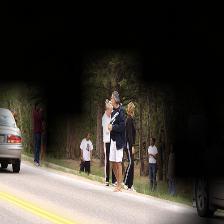}
\\
\includegraphics[width=\linewidth, height=25mm]{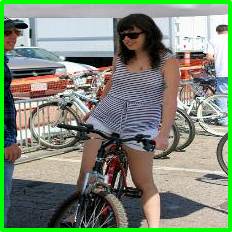} &
\includegraphics[width=\linewidth, height=25mm]{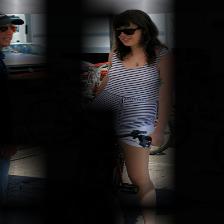} &
\includegraphics[width=\linewidth, height=25mm]{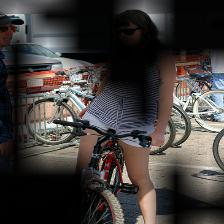} &
&
\includegraphics[width=\linewidth, height=25mm]{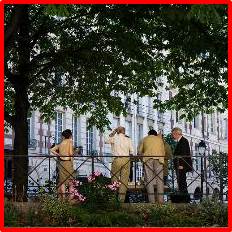} &
\includegraphics[width=\linewidth, height=25mm]{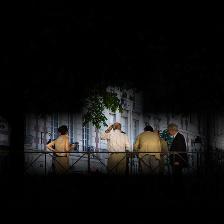} &
\includegraphics[width=\linewidth, height=25mm]{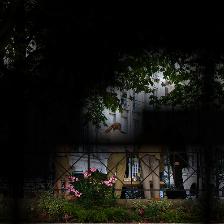}
\\
\end{tabular}
\vspace{-2mm}
\captionof{figure}{
Qualitative results from text-to-image retrieval with attention visualization.
For each example, the query sentence and the top two retrieved images are shown from top to bottom; the original sentence (image), the first and second attention maps are shown from left to right.
Green and red boxes indicate ground-truth and non ground-truth images, respectively.
}
\label{fig:txt2img}
\end{table*}

\subsection{Evaluation on Visual Question Answering}

\subsubsection{Dataset and Evaluation Metric}
We evaluate the r-DAN on the Visual Question Answering (VQA) dataset~\cite{antol2015vqa}, which contains approximately 200K real images from MSCOCO dataset~\cite{lin2014microsoft}.
Each image is associated with three questions, and each question is labeled with ten answers by human annotators.
The dataset is typically divided into four splits: \emph{train} (80K images), \emph{val} (40K images), \emph{test-dev} (20K images), and \emph{test-std} (20K images).
We train our model using \emph{train} and \emph{val}, validate with \emph{test-dev}, and evaluate on \emph{test-std}.
There are two forms of tasks, open-ended and multiple-choice, which require to answer each question without and with a set of candidate answers, respectively.
For both tasks, we follow the evaluation metric used in \cite{antol2015vqa} as
\begin{align}
	\operatorname{Acc}(\hat{a}) = \min \left\{ \frac{\text{\#humans that labeled }\hat{a}}{3}, 1 \right\},
\label{eq:acc}
\end{align}
where $\hat{a}$ is a predicted answer.

\subsubsection{Results and Analysis}
The performance of r-DAN compared with state-of-the-art VQA systems is presented in Table~\ref{tab:vqa}, where our method achieves the best performance in both open-ended and multiple-choice tasks.
For fair evaluation, we compare single-model accuracies obtained without data augmentation, even though \cite{fukui2016multimodal} reported better performance using model ensembles and additional training data.
Figure~\ref{fig:vqa_qual} describes the qualitative results from our approach with visualization of the attention weights.
Our method produces correct answers to challenging problems which require fine-grained reasoning, as well as successfully attends to the specific regions and words which facilitate answering the questions.
Specifically, the first and fourth examples in Figure~\ref{fig:vqa_qual} illustrate that the r-DAN moves its visual attention to the proper regions indicated by the attended words, while the second and third examples show that it moves its textual attention to divide a complex task into sequential subtasks---finding target objects and extracting certain attributes.


\subsection{Evaluation on Image-Text Matching}

\subsubsection{Dataset and Evaluation Metric}
We employ the Flickr30K dataset~\cite{young2014image} to evaluate the m-DAN for multimodal matching.
It consists of 31,783 real images with five descriptive sentences for each, and we follow the public splits by \cite{mao2015deep}: 29,783 training, 1,000 validation and 1,000 test images.
We report the performance of m-DAN in bidirectional image and sentence retrieval using the same metrics as previous work~\cite{yan2015deep,ma2015multimodal,mao2015deep,wang2016learning}.
Recall@K (K=1, 5, 10) represents the percentage of the queries where at least one ground-truth is retrieved among the top K results and MR measures the median rank of the top-ranked ground-truth.

\subsubsection{Results and Analysis}
Table~\ref{tab:matching} presents the quantitative results on the Flickr30K dataset, where the proposed method outperforms other recent approaches in all measures.
The qualitative results from image-to-text and text-to-image retrieval are also illustrated in Figure~\ref{fig:img2txt} and Figure~\ref{fig:txt2img}, respectively, with visualization of attention outputs.
At each step of attention, the m-DAN effectively discovers the essential semantics appearing in both modalities.
It tends to capture the main subjects (\eg~{\tt woman}, {\tt boy}, {\tt people}, etc.) at the first step, and figure out relevant objects, backgrounds or actions (\eg~{\tt computer}, {\tt scaffolding}, {\tt sweeps}, etc.) at the second step. 
Note that this property solely comes from the training stage where visual and textual attention models are jointly learned, while images and sentences are processed independently at inference time.

\section{Conclusion}
We propose Dual Attention Networks (DANs) to bridge visual and textual attention mechanisms.
We present two architectures of DANs for multimodal reasoning and matching.
The first model infers the answers collaboratively from images and sentences, while the other one embeds them into a common space by capturing their shared semantics.
These models demonstrate the state-of-the-art performance in VQA and image-text matching, showing their effectiveness in extracting essential information via the dual attention mechanism.
The proposed framework can be potentially generalized to various tasks at the intersection of vision and language, such as image captioning, visual grounding, video question answering, etc. 

{\small
\bibliographystyle{ieee}
\bibliography{egbib}
}

\end{document}